\documentclass[runningheads]{llncs}
\usepackage[T1]{fontenc}
\usepackage{graphicx}
\usepackage{subcaption,bbm}
\usepackage{booktabs,array,graphicx}
\usepackage{multirow}
\usepackage{algorithm}
\usepackage{algpseudocode}

\newcommand{\ind}[1]{\mathbbm{1}_{(#1)}}

\usepackage{dcolumn}
\newcolumntype{d}[1]{D{.}{.}{#1}}
\newcolumntype{B}[3]{>{\boldmath\DC@{#1}{#2}{#3}}c<{\DC@end}}

\usepackage[detect-all]{siunitx}
\usepackage{textcomp}

\usepackage[acronym]{glossaries}
\newacronym{ice}{ICE}{Intracardiac Echocardiography}
\newacronym{af}{AFib}{Atrial Fibrillation}
\newacronym{la}{LA}{Left Atrium}
\newacronym{ct}{CT}{Computed Tomography}
\newacronym{cas}{CAS}{Clinical Application Specialist}
\newacronym{ce}{CE}{Cross-Entropy}

\newcommand{\ice}{\acrshort{ice} }

\newcommand\footnotenomarker[1]{%
  \begingroup
  \renewcommand\thefootnote{}\footnote{#1}%
  \addtocounter{footnote}{-1}%
  \endgroup
}

\begin{document}
\title{From Sparse to Precise: A Practical Editing Approach for Intracardiac Echocardiography Segmentation}
\titlerunning{A Practical Editing Approach for ICE Segmentation}
\author{Ahmed H. Shahin\inst{1,2}\thanks{Corresponding author} \and
Yan Zhuang\inst{1,3} \and
Noha El-Zehiry\inst{1,4}}
\index{Shahin, Ahmed H.}
\index{Zhuang, Yan}
\index{El-Zehiry, Noha}

\authorrunning{A. Shahin et al.}
\institute{Siemens Healthineers, New Jersey, USA \and University College London, London, UK \and National Institutes of Health Clinical Center, Maryland, USA \and Wipro, New Jersey, USA\\
\email{ahmedhshahen@gmail.com}
}
\maketitle              %
\footnotenomarker{Work done while the authors were employed by Siemens Healthineers.}
\footnotenomarker{Disclaimer: The concepts and information presented in this paper are based on research results that are not commercially available. Future commercial availability cannot be guaranteed.}
\begin{abstract}
    Accurate and safe catheter ablation procedures for atrial fibrillation require precise segmentation of cardiac structures in Intracardiac Echocardiography (ICE) imaging. Prior studies have suggested methods that employ 3D geometry information from the ICE transducer to create a sparse ICE volume by placing 2D frames in a 3D grid, enabling the training of 3D segmentation models. However, the resulting 3D masks from these models can be inaccurate and may lead to serious clinical complications due to the sparse sampling in ICE data, frames misalignment, and cardiac motion. To address this issue, we propose an interactive editing framework that allows users to edit segmentation output by drawing scribbles on a 2D frame.  The user interaction is mapped to the 3D grid and utilized to execute an editing step that modifies the segmentation in the vicinity of the interaction while preserving the previous segmentation away from the interaction. Furthermore, our framework accommodates multiple edits to the segmentation output in a sequential manner without compromising previous edits. This paper presents a novel loss function and a novel evaluation metric specifically designed for editing. Cross-validation and testing results indicate that, in terms of segmentation quality and following user input, our proposed loss function outperforms standard losses and training strategies. We demonstrate quantitatively and qualitatively that subsequent edits do not compromise previous edits when using our method, as opposed to standard segmentation losses. Our approach improves segmentation accuracy while avoiding undesired changes away from user interactions and without compromising the quality of previously edited regions, leading to better patient outcomes.
\end{abstract}

\setcounter{footnote}{0} 
\section{Introduction}
\acrfull{af} is a prevalent cardiac arrhythmia affecting over 45 million individuals worldwide as of 2016~\cite{kornej2020}.
Catheter ablation, which involves the elimination of affected cardiac tissue, is a widely used treatment for \acrshort{af}. To ensure procedural safety and minimize harm to healthy tissue, \acrfull{ice} imaging is utilized to guide the intervention.

\acrlong{ice} imaging utilizes an ultrasound probe attached to a catheter and inserted into the heart to obtain real-time images of its internal structures. In ablation procedures for \acrfull{la} \acrshort{af} treatment, the \acrshort{ice} ultrasound catheter is inserted in the right atrium to image the left atrial structures. The catheter is rotated clockwise to capture image frames that show the \acrshort{la} body, the \acrshort{la} appendage and the pulmonary veins~\cite{russo2013}. Unlike other imaging modalities, such as transesophageal echocardiography, \ice imaging does not require general anesthesia~\cite{enriquez2018}. Therefore, it is a safer and more convenient option for cardiac interventions using ultrasound imaging.

The precise segmentation of cardiac structures, particularly the \acrshort{la}, is crucial for the success and safety of catheter ablation. However, segmentation of the \acrshort{la} is challenging due to the constrained spatial resolution of 2D \ice images and the manual manipulation of the \ice transducer. Additionally, the sparse sampling of \ice frames makes it difficult to train automatic segmentation models. Consequently, there is a persistent need to develop interactive editing tools to help experts modify the automatic segmentation to reach clinically satisfactory accuracy.

\begin{figure}[!t]
    \centering
    \begin{subfigure}[b]{0.24\textwidth}
        \centering
        \includegraphics[width=\textwidth]{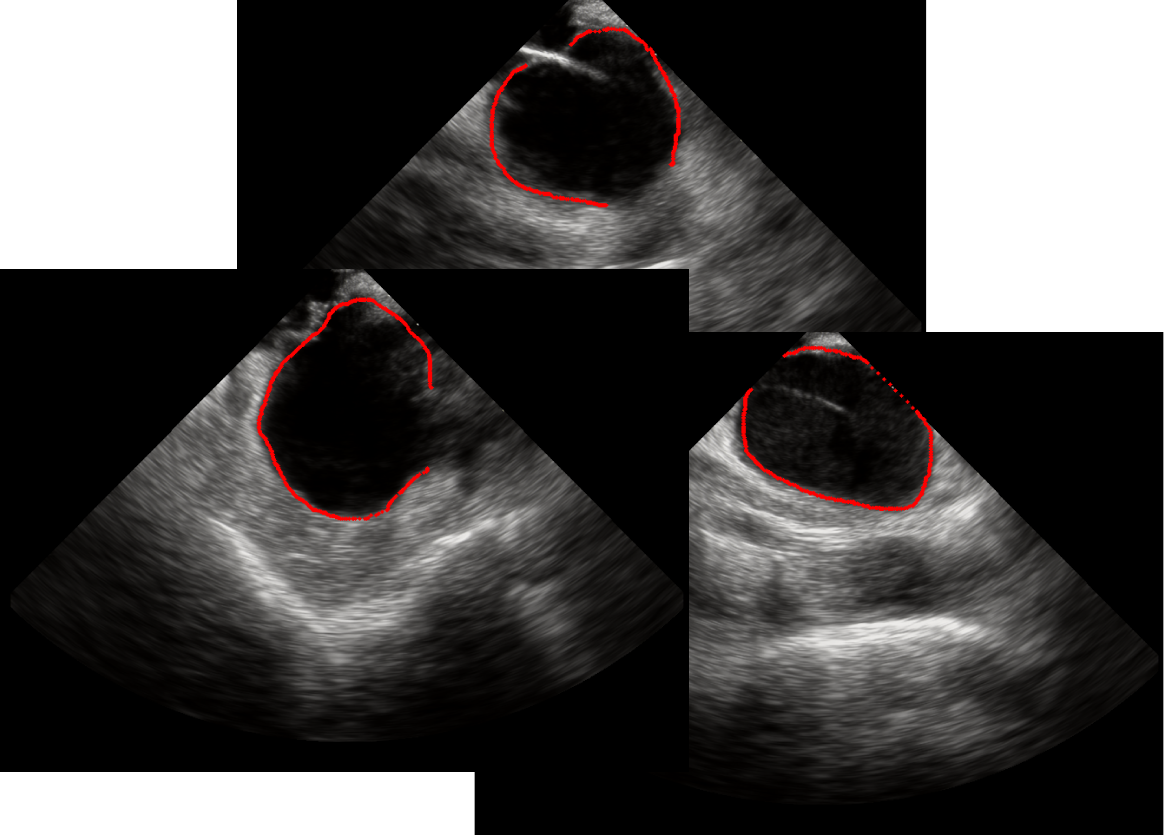}
        \caption{}
        \label{fig:ice2d}
    \end{subfigure}
    \begin{subfigure}[b]{0.24\textwidth}
        \centering
        \includegraphics[width=0.75\textwidth]{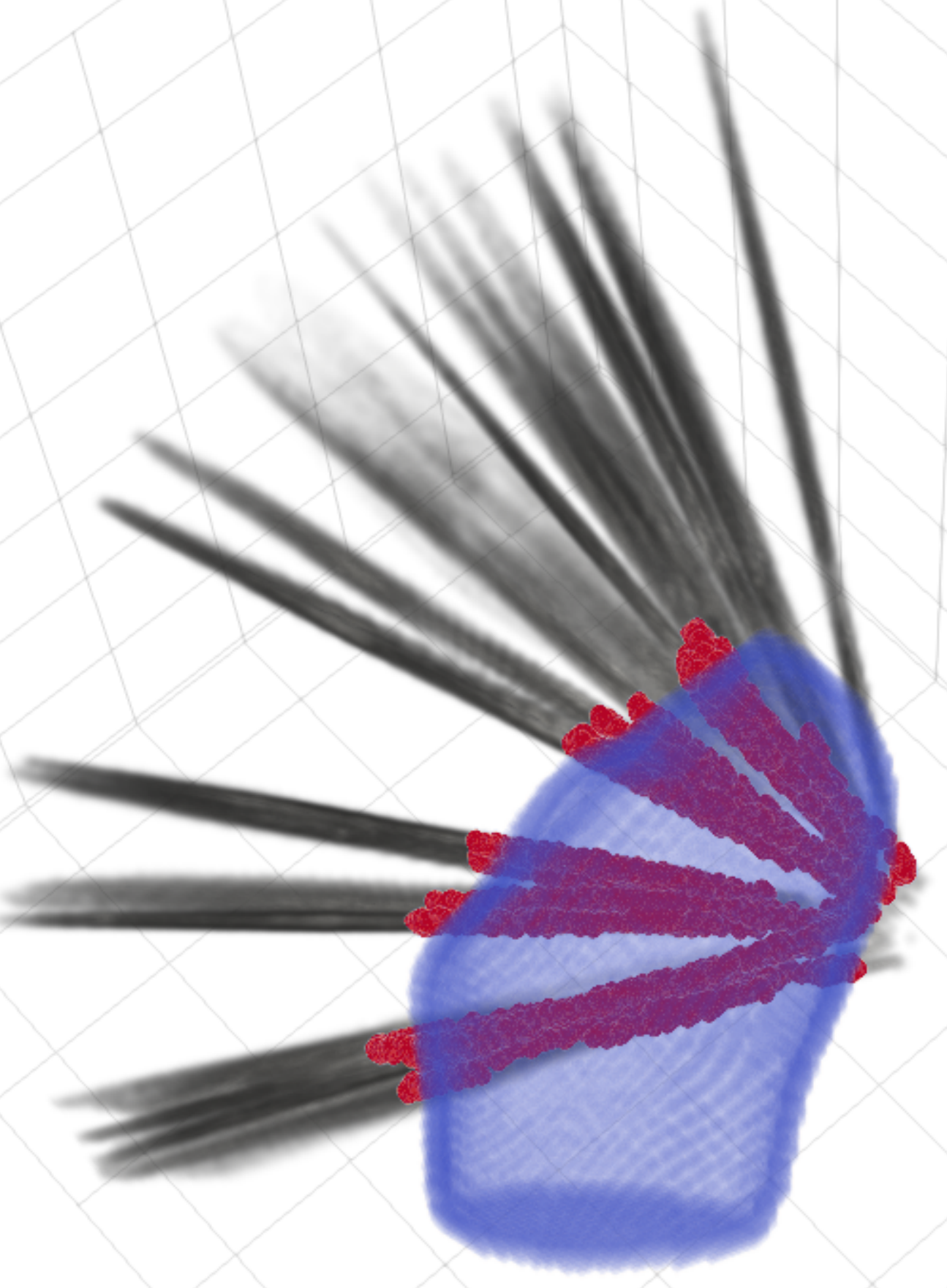}
        \caption{}
        \label{fig:ice3d}
    \end{subfigure}
    \begin{subfigure}[b]{0.24\textwidth}
        \centering
        \includegraphics[width=0.75\textwidth]{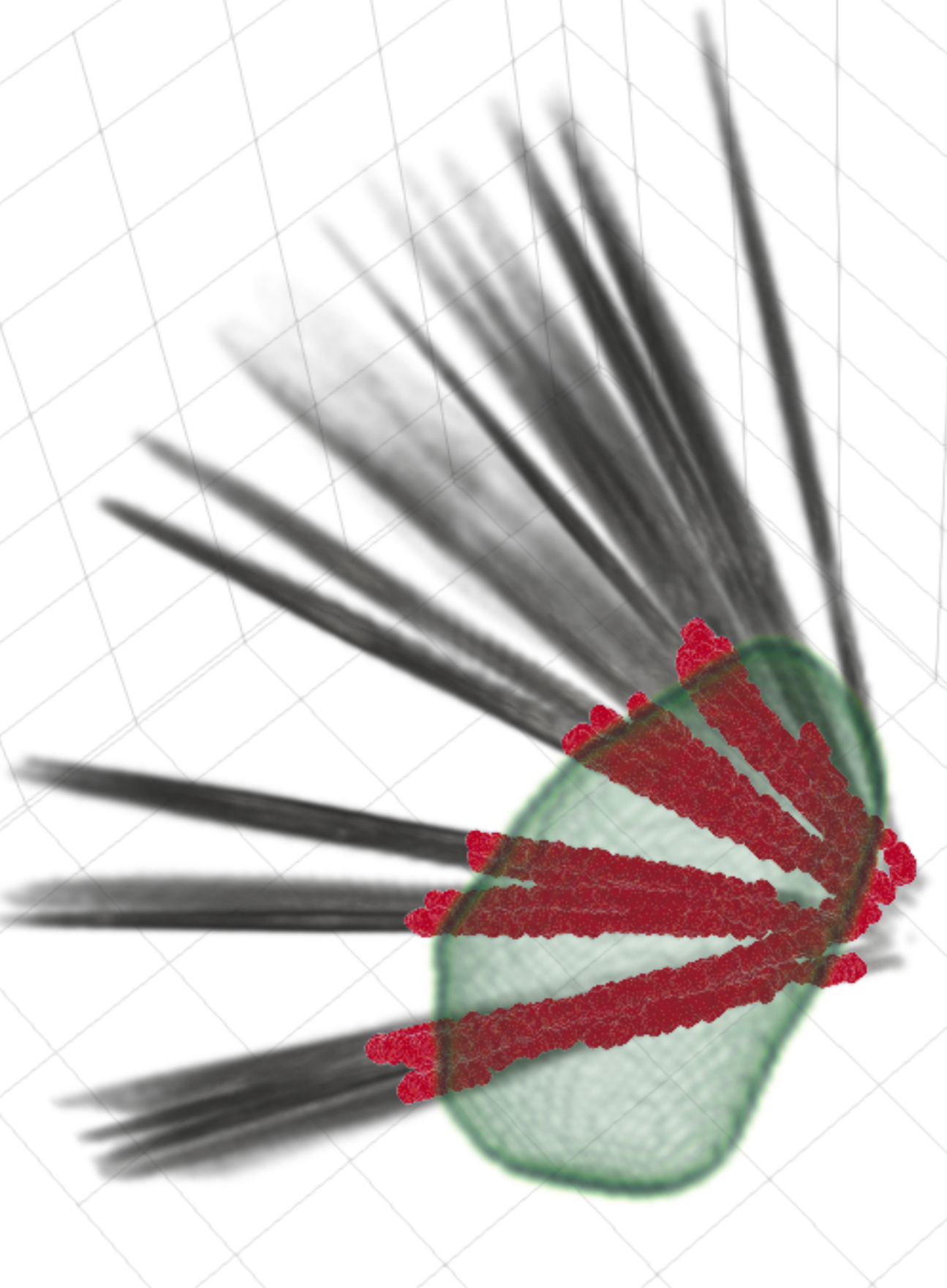}
        \caption{}
        \label{fig:ice_pred}
    \end{subfigure}
    \begin{subfigure}[b]{0.24\textwidth}
        \centering
        \includegraphics[width=\textwidth]{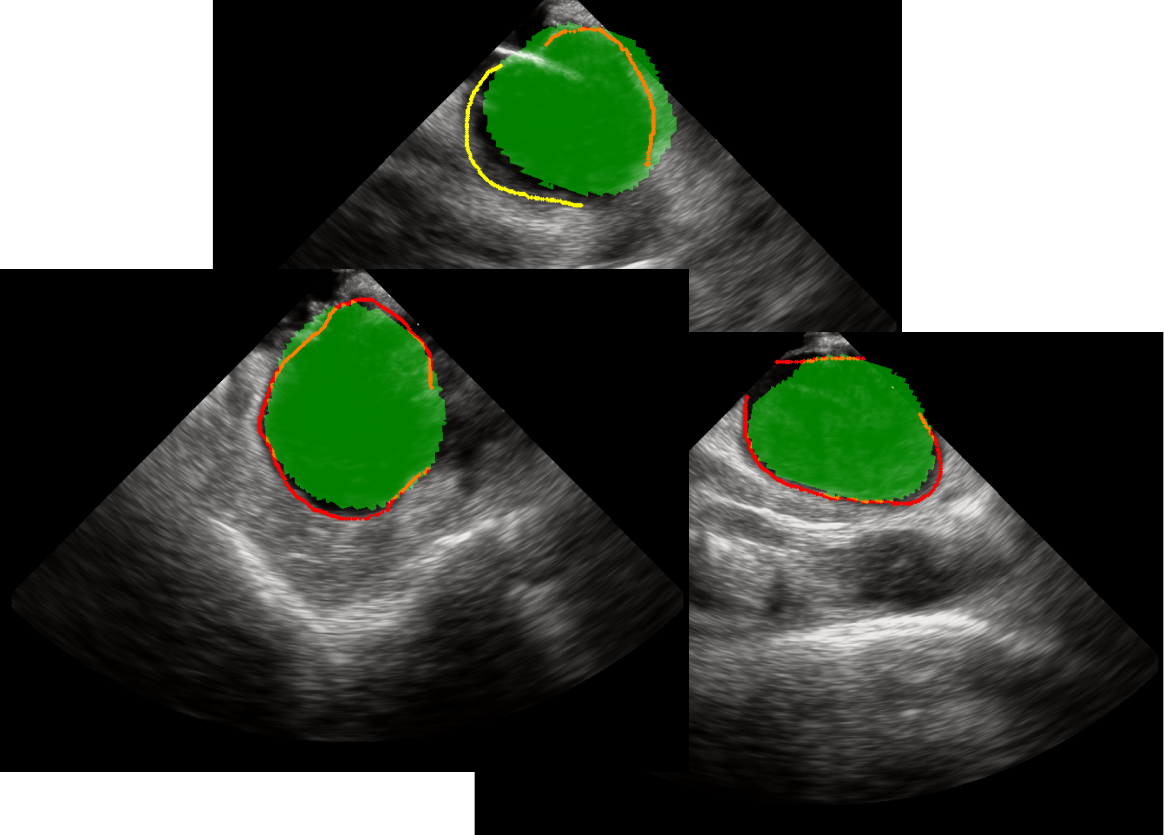}
        \caption{}
        \label{fig:ice_edit}
    \end{subfigure}
    \caption{Volumetric segmentation of \acrshort{ice} data. (a) 2D \acrshort{ice} frames with \acrshort{cas} contours outlining \acrshort{la} boundaries. (b) 2D frames (black shades) and \acrshort{cas} contours projected onto a 3D grid. The blue mesh represents the 3D segmentation mask obtained by deforming a \acrshort{ct} template to fit the contours as closely as possible \cite{liao2018}. Note the sparsity of frames. (c) Predicted 3D segmentation mask generated by a model trained with masks from (b). (d) Predicted mask projected onto 2D (green) and compared with the original \acrshort{cas} contours. Note the misalignment between the mask and \acrshort{cas} contours in some frames. Yellow indicates an example of a user-corrective edit.}
\end{figure}

During a typical \ice imaging scan, a series of sparse 2D \ice frames is captured and a \acrfull{cas} annotates the boundaries of the desired cardiac structure in each frame\footnote{Annotations typically take the form of contours instead of masks, as the structures being segmented appear with open boundaries in the frames.} (Fig. \ref{fig:ice2d}). To construct dense 3D masks for training segmentation models, Liao et al. utilized the 3D geometry information from the \ice transducer, to project the frames and their annotations onto a 3D grid~\cite{liao2018}. They deformed a 3D template of the \acrshort{la} computed from 414 CT scans to align as closely as possible with the \acrshort{cas} contours, producing a 3D mesh to train a segmentation model~\cite{liao2018}. However, the resulting mesh may not perfectly align with the original \acrshort{cas} contours due to factors such as frames misalignment and cardiac motion (Fig. \ref{fig:ice3d}). Consequently, models trained with such 3D mesh as ground truth do not produce accurate enough segmentation results, which can lead to serious complications (Fig. \ref{fig:ice_pred}).

A natural remedy is to allow clinicians to edit the segmentation output and create a model that incorporates and follows these edits. In the case of \ice data, the user interacts with the segmentation output by drawing a scribble on one of the 2D frames (Fig. \ref{fig:ice_edit}). Ideally, the user interaction should influence the segmentation in the neighboring frames while preserving the original segmentation in the rest of the volume. Moreover, the user may make multiple edits to the segmentation output, which must be incorporated in a sequential manner without compromising the previous edits.%

In this paper, we present a novel interactive editing framework for the \ice data. This is the first study to address the specific challenges of interactive editing with \ice data. Most of the editing literature treats editing as an interactive segmentation problem and does not provide a clear distinction between interactive segmentation and interactive editing. We provide a novel method that is specifically designed for editing. The novelty of our approach is two-fold: 1) We introduce an editing-specific novel loss function that guides the model to incorporate user edits \textit{while preserving the original segmentation in unedited areas}. 2) We present a novel evaluation metric that best reflects the editing formulation. Comprehensive evaluations of the proposed method on \ice data demonstrate that the presented loss function achieves superior performance compared to traditional interactive segmentation losses and training strategies, as evidenced by the experimental data.

\section{Interactive Editing of \ice Data}
\subsection{Problem Definition}
\label{sec:setup}
The user is presented first with an \ice volume, $x \in R^{H \times W \times D}$, and its initial imperfect segmentation, $y_{\text{init}} \in R^{H \times W \times D}$, where $H$, $W$ and $D$ are the dimensions of the volume. To correct inaccuracies in the segmentation, the user draws a scribble on one of the 2D \ice frames. Our goal is to use this 2D interaction to provide a 3D correction to $y_{\text{init}}$ in the vicinity of the user interaction. We project the user interaction from 2D to 3D and encode it as a 3D Gaussian heatmap, $u \in R^{H \times W \times D}$, centered on the scribble with a standard deviation of $\sigma_{enc}$~\cite{maninis2018}. The user iteratively interacts with the output until they are satisfied with the quality of the segmentation.

We train an editing model $f$ to predict the corrected segmentation output $\hat{y}^t \in R^{H\times W\times D}$ given $x$, $y_{\text{init}}^t$, and $u^t$, where $t$ is the iteration number. The goal is for $\hat{y}^t$ to accurately reflect the user's correction near their interaction while preserving the initial segmentation elsewhere. Since $y_{\text{init}}^{t+1} \equiv \hat{y}^t$, subsequent user inputs $u^{\{t+1, \dots, T\}}$ should not corrupt previous corrections $u^{\{0, \dots, t\}}$.

\begin{figure}[!t]
    \centering
    \includegraphics[width=\textwidth]{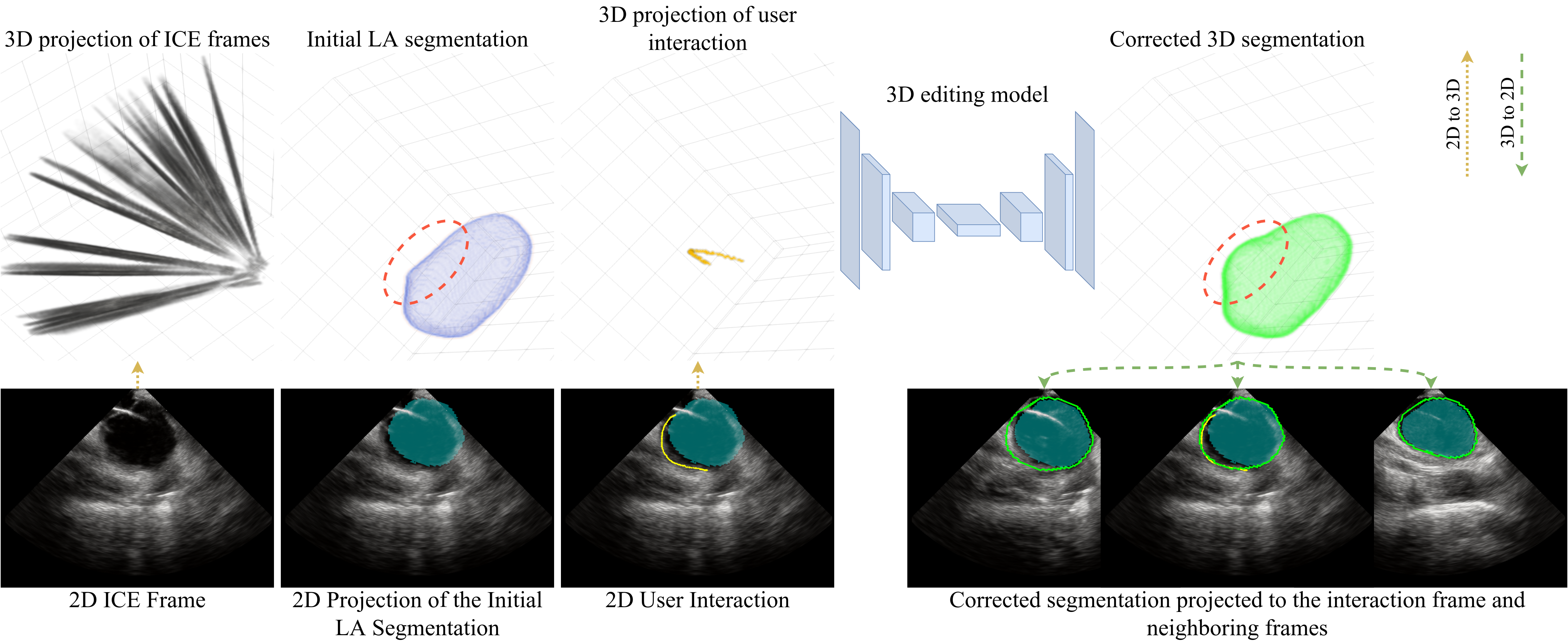}
    \caption{The proposed interactive editing framework involves user interaction with the segmentation output by drawing a scribble on one of the 2D frames. The editing model is trained to incorporate user interaction while preserving the initial segmentation in unedited areas. Cyan shade: initial segmentation. Green contour: corrected segmentation. Yellow contour: user interaction.}
\end{figure}

\subsection{Loss Function}
\label{sec:loss}
Most interactive segmentation methods aim to incorporate user guidance to enhance the overall segmentation~\cite{reuben2021,khan2019,maninis2018}. However, in our scenario, this approach may undesirably modify previously edited areas and may not align with clinical expectations since the user has corrected these areas, and the changes are unexpected. To address the former issue, Bredell et al. proposed an iterative training strategy in which user edits are synthesized and accumulated over a fixed number of steps with every training iteration~\cite{bredell2018}. However, this approach comes with a significant increase in training time and does not explicitly instruct the model to preserve regions away from the user input.

We propose an \textbf{editing-specific} loss function $\mathcal{L}$ that encourages the model to preserve the initial segmentation while incorporating user input. The proposed loss function incentivizes the model to match the prediction $\hat{y}$ with the ground truth $y$ in the vicinity of the user interaction. In regions further away from the user interaction, the loss function encourages the model to match the initial segmentation $y_{\text{init}}$, instead. Here, $y$ represents the 3D mesh, which is created by deforming a CT template to align with the CAS contours $y_{cas}$ \cite{liao2018}. Meanwhile, $y_{init}$ denotes the output of a segmentation model that has been trained on $y$.

We define the vicinity of the user interaction as a 3D Gaussian heatmap, $A \in R^{H \times W \times D}$, centered on the scribble with a standard deviation of $\sigma_{edit}$. Correspondingly, the regions far from the interaction are defined as $\bar{A}=1-A$. The loss function is defined as the sum of the weighted cross entropy losses $\mathcal{L}_{\text{edit}}$ and $\mathcal{L}_{\text{preserve}}$ w.r.t $y$ and $y_{\text{init}}$, respectively, as follows
\begin{equation}
    \mathcal{L} = \mathcal{L}_{\text{edit}} + \mathcal{L}_{\text{preserve}}
\end{equation}
where
\begin{align}
    \mathcal{L}_{\text{edit}} &= -\sum_{i=1}^{H} \sum_{j=1}^{W} \sum_{k=1}^{D} A_{i,j,k} \left[ y_{i,j,k} \log \hat{y}_{i,j,k} + (1-y_{i,j,k}) \log (1-\hat{y}_{i,j,k}) \right]\\
    \mathcal{L}_{\text{preserve}} &= -\sum_{i=1}^{H} \sum_{j=1}^{W} \sum_{k=1}^{D} \bar{A}_{i,j,k} \left[ y_{{init}_{i,j,k}} \log \hat{y}_{i,j,k} + (1-y_{{init}_{i,j,k}}) \log (1-\hat{y}_{i,j,k}) \right]
\end{align}
The Gaussian heatmaps facilitate a gradual transition between the edited and unedited areas, resulting in a smooth boundary between the two regions.

\subsection{Evaluation Metric}
\label{sec:metric}
The evaluation of segmentation quality typically involves metrics such as the Dice coefficient and the Jaccard index, which are defined for binary masks, or distance-based metrics, which are defined for contours~\cite{taha2015}. In our scenario, where the ground truth is \acrshort{cas} contours, we use distance-based metrics\footnote{Contours are inferred from the predicted mask $\hat{y}$.}. However, standard utilization of these metrics computes the distance between the predicted and ground truth contours, which misleadingly incentivizes alignment with the ground truth contours in \textbf{all regions}. This approach incentivizes changes in the unedited regions, which is undesirable from a user perspective, as users want to see changes only in the vicinity of their edit. Additionally, this approach incentivizes the corruption of previous edits.

We propose a novel editing-specific evaluation metric that assesses how well the prediction $\hat{y}$ matches the \acrshort{cas} contours $y_{\text{cas}}$ in the vicinity of the user interaction, and the initial segmentation $y_{\text{init}}$ in the regions far from the interaction.
\begin{equation}
    \mathcal{D} = \mathcal{D}_{\text{edit}} + \mathcal{D}_{\text{preserve}}
\end{equation}
where, $\forall (i,j,k) \in \{1, \dots, H\} \times \{1, \dots, W\} \times \{1, \dots, D\}$, $\mathcal{D}_{\text{edit}}$ is the distance from $y_{\text{cas}}$ to $\hat{y}$ in the vicinity of the user edit, as follows
\begin{equation}
    \mathcal{D}_{\text{edit}} = 
        \ind{y_{\text{cas}_{i,j,k}}=1} \cdot A_{i,j,k} \cdot d(y_{\text{cas}_{i,j,k}}, \hat{y})
\end{equation}
where $d$ is the minimum Manhattan distance from $y_{\text{cas}_{i,j,k}}$ to any point on $\hat{y}$. For $\mathcal{D}_{\text{preserve}}$, we compute the average symmetric distance between $y_{\text{init}}$ and $\hat{y}$, since the two contours are of comparable length. The average symmetric distance is defined as the average of the minimum Manhattan distance from each point on $y_{\text{init}}$ contour to $\hat{y}$ contour and vice versa, as follows
\begin{equation}
    \mathcal{D_{\text{preserve}}} = \frac{\bar{A}}{2} \cdot \left[ \ind{y_{\text{init}_{i,j,k}}=1} \cdot d(y_{\text{init}_{i,j,k}}, \hat{y}) + \ind{\hat{y}_{i,j,k}=1} \cdot d(\hat{y}_{i,j,k}, y_{\text{init}}) \right]
\end{equation}
The resulting $\mathcal{D}$ represents a distance map $\in R^{H \times W \times D}$ with defined values only on the contours $y_{\text{cas}}, y_{\text{init}}, \hat{y}$. Statistics such as the $95^{\text{th}}$ percentile and mean can be computed on the corresponding values of these contours on the distance map.

\section{Experiments}
\subsection{Dataset}
\label{sec:dataset}
Our dataset comprises \ice scans for 712 patients, each with their \acrshort{la} \acrshort{cas} contours $y_{cas}$ and the corresponding 3D meshes $y$ generated by~\cite{liao2018}. Scans have an average of 28 2D frames. Using the 3D geometry information, frames are projected to a 3D grid with a resolution of $128\times128\times128$ and voxel spacing of $1.1024\times1.1024\times1.1024$ mm. We performed five-fold cross-validation on $85\%$ of the dataset (605 patients) and used the remaining $15\%$ (107 patients) for testing.

\subsection{Implementation Details}
\label{sec:implementation}
To obtain the initial imperfect segmentation $y_{\text{init}}$, a U-Net model~\cite{ronneberger2015} is trained on the 3D meshes $y$ using a \acrfull{ce} loss. The same U-Net architecture is used for the editing model. The encoding block consists of two 3D convolutional layers followed by a max pooling layer. Each convolutional layer is followed by batch normalization and ReLU non-linearity layers~\cite{ioffe2015}. The number of filters in the segmentation model convolutional layers are 16, 32, 64, and 128 for each encoding block, and half of them for the editing model. The decoder follows a similar architecture.

The input of the editing model consists of three channels: the input \ice volume $x$, the initial segmentation $y_{\text{init}}$, and the user input $u$. During training, the user interaction is synthesized on the frame with maximum error between $y_{\text{init}}$ and $y$.\footnote{We do not utilize the \acrshort{cas} contours during training and only use them for testing because the \acrshort{cas} contours do not align with the segmentation meshes $y$.} The region of maximum error is selected and a scribble is drawn on the boundary of the ground truth in that region to simulate the user interaction. During testing, the real contours of the \acrshort{cas} are used and the contour with the maximum distance from the predicted segmentation is chosen as the user interaction. The values of $\sigma_{enc}$ and $\sigma_{edit}$ are set to 20, chosen by cross-validation.  Adam optimizer is used with a learning rate of 0.005 and a batch size of 4 to train the editing model for 100 epochs~\cite{kingma2014}.

\subsection{Results}
\label{sec:results}
We use the editing evaluation metric $\mathcal{D}$ (Sec. \ref{sec:metric}) for the evaluation of the different methods. For better interpretability of the results, we report the overall error, the error near the user input, and the error far from the user input. We define near and far regions by thresholding the Gaussian heatmap $A$ at $0.5$.

\begin{table}[!t]
    \centering
    \caption{Results on Cross-Validation (CV) and test set. We use the editing evaluation metric $\mathcal{D}$ and report the $95^{\text{th}}$ percentile of the overall editing error, the error near the user input, and far from the user input (mm). The near and far regions are defined by thresholding $A$ at $0.5$. For the CV results, we report the mean and standard deviation over the five folds. The statistical significance is computed for the difference with InterCNN. $^\dagger$: p-value $<0.01$, $^\ddagger$: p-value $<0.001$.}
    \label{tab:results_single}
    \resizebox{\textwidth}{!}{
    \begin{tabular}{@{} m{1.9cm} r@{\text{} $\pm$ \text{}}l r@{\text{} $\pm$ \text{}}l r@{\text{} $\pm$ \text{}}l r@{.}l r@{.}l r@{.}l @{}}
    \toprule
    \multirow{2}{*}{Method} & \multicolumn{6}{c}{CV} & \multicolumn{6}{c}{Test} \\ \cmidrule(l{3pt}r{3pt}){2-7} \cmidrule(l{3pt}r{3pt}){8-13} & \multicolumn{2}{c}{Overall $\downarrow$} & \multicolumn{2}{c}{Near $\downarrow$} & \multicolumn{2}{c}{Far $\downarrow$} & \multicolumn{2}{c}{Overall $\downarrow$} & \multicolumn{2}{c}{Near $\downarrow$} & \multicolumn{2}{c}{Far $\downarrow$} \\ \midrule
    No Editing & 3.962 & 0.148 &   \multicolumn{2}{c}{-} & \multicolumn{2}{c}{-} & 4&126 & \multicolumn{2}{c}{-} & \multicolumn{2}{c}{-} \\
    \acrshort{ce} Loss & 1.164 & 0.094 & 0.577 & 0.024 & 0.849 & 0.105 & 1&389 & 0&6 & 1&073 \\
    Dice Loss  & 1.188 & 0.173 & 0.57 & 0.089  & 0.892 & 0.155 & 1&039 & \textbf{0}&\textbf{46} & 0&818 \\
    InterCNN   & 0.945 & 0.049 & \textbf{0.517} & \textbf{0.052} & 0.561 & 0.006 & 0&94 & 0&509 & 0&569 \\
    Editing Loss & \textbf{0.809} & \textbf{0.05}$^\ddagger$ & 0.621 & 0.042 & \textbf{0.182} & \textbf{0.01}$^\ddagger$ & \textbf{0}&\textbf{844}$^\dagger$ & 0&662 & \textbf{0}&\textbf{184}$^\ddagger$ \\ \bottomrule
    \end{tabular}
    }
\end{table}

We evaluate our loss (editing loss) against the following baselines: (1) \textbf{No Editing}: the initial segmentation $y_{\text{init}}$ is used as the final segmentation $\hat{y}$, and the overall error in this case is the distance from the \acrshort{cas} contours to $y_{\text{init}}$. This should serve as an upper bound for error. (2) \textbf{\acrshort{ce} Loss}: an editing model trained using the standard \acrshort{ce} segmentation loss w.r.t $y$. (3) \textbf{Dice Loss}~\cite{Milletari2016}: an editing model trained using Dice segmentation loss w.r.t $y$. (4) \textbf{InterCNN}~\cite{bredell2018}: for every training sample, simulated user edits based on the prediction are accumulated with any previous edits and re-input to the model for 10 iterations, trained using \acrshort{ce} loss. We report the results after a single edit (the furthest \acrshort{cas} contour from $\hat{y}$) in Table \ref{tab:results_single}. A single training epoch takes $\approx 3$ minutes for all models except InterCNN, which takes $\approx 14$ minutes, on a single NVIDIA Tesla V100 GPU. The inference time through our model is $\approx 20$ milliseconds per volume.

Our results demonstrate that the proposed loss outperforms all baselines in terms of overall error. Although all the editing methods exhibit comparable performance in the near region, in the far region where the error is calculated relative to $y_{\text{init}}$, our proposed loss outperforms all the baselines by a significant margin. This can be attributed to the fact that the baselines are trained using loss functions which aim to match the ground truth globally, resulting in deviations from the initial segmentation in the far region. In contrast, our loss takes into account user input in its vicinity and maintains the initial segmentation elsewhere. 

\subsubsection{Sequential Editing}
\label{sec:seq_editing} 
\begin{figure}[!t]
    \centering
    \includegraphics[width=0.36\textwidth]{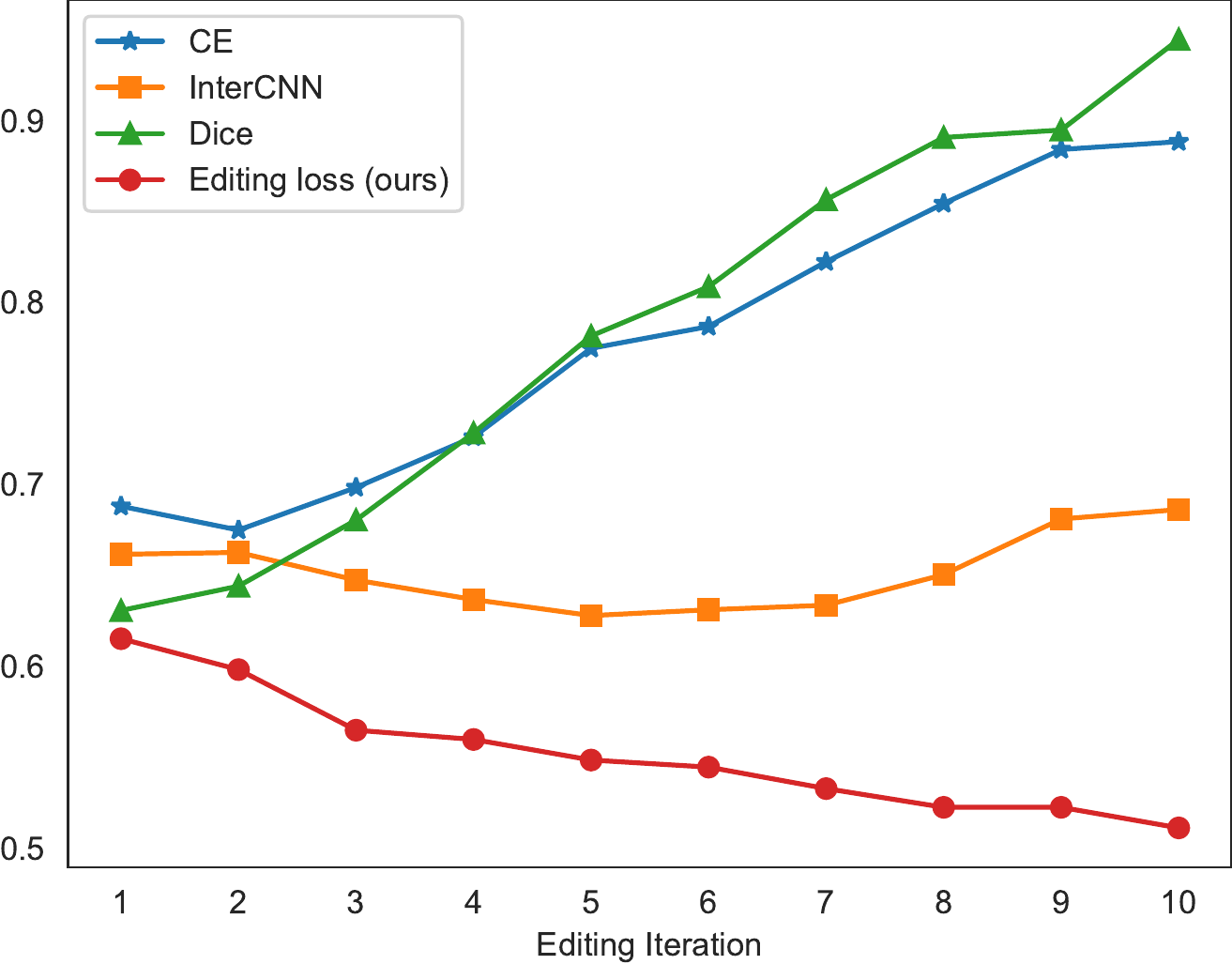}
    \caption{$95^{\text{th}}$ percentile of the distance from the \acrshort{cas} contours to the prediction.}
    \label{fig:seq_editing}
\end{figure}

We also investigate the scenario in which the user iteratively performs edits on the segmentation multiple times. We utilized the same models that were used in the single edit experiment and simulated 10 editing iterations. At each iteration, we selected the furthest \acrshort{cas} contour from $\hat{y}$, ensuring that the same edit was not repeated twice. For the interCNN model, we aggregated the previous edits and input them into the model, whereas for all other models, we input a single edit per iteration. We assessed the impact of the number of edits on the overall error. In Fig. \ref{fig:seq_editing}, we calculated the distance from all the \acrshort{cas} contours to the predicted segmentation and observed that the editing loss model improved with more edits. In contrast, the CE and Dice losses degraded with more edits due to compromising the previous corrections, while InterCNN had only marginal improvements.

\begin{figure}[!h]
    \centering
    \includegraphics[width=0.74\textwidth]{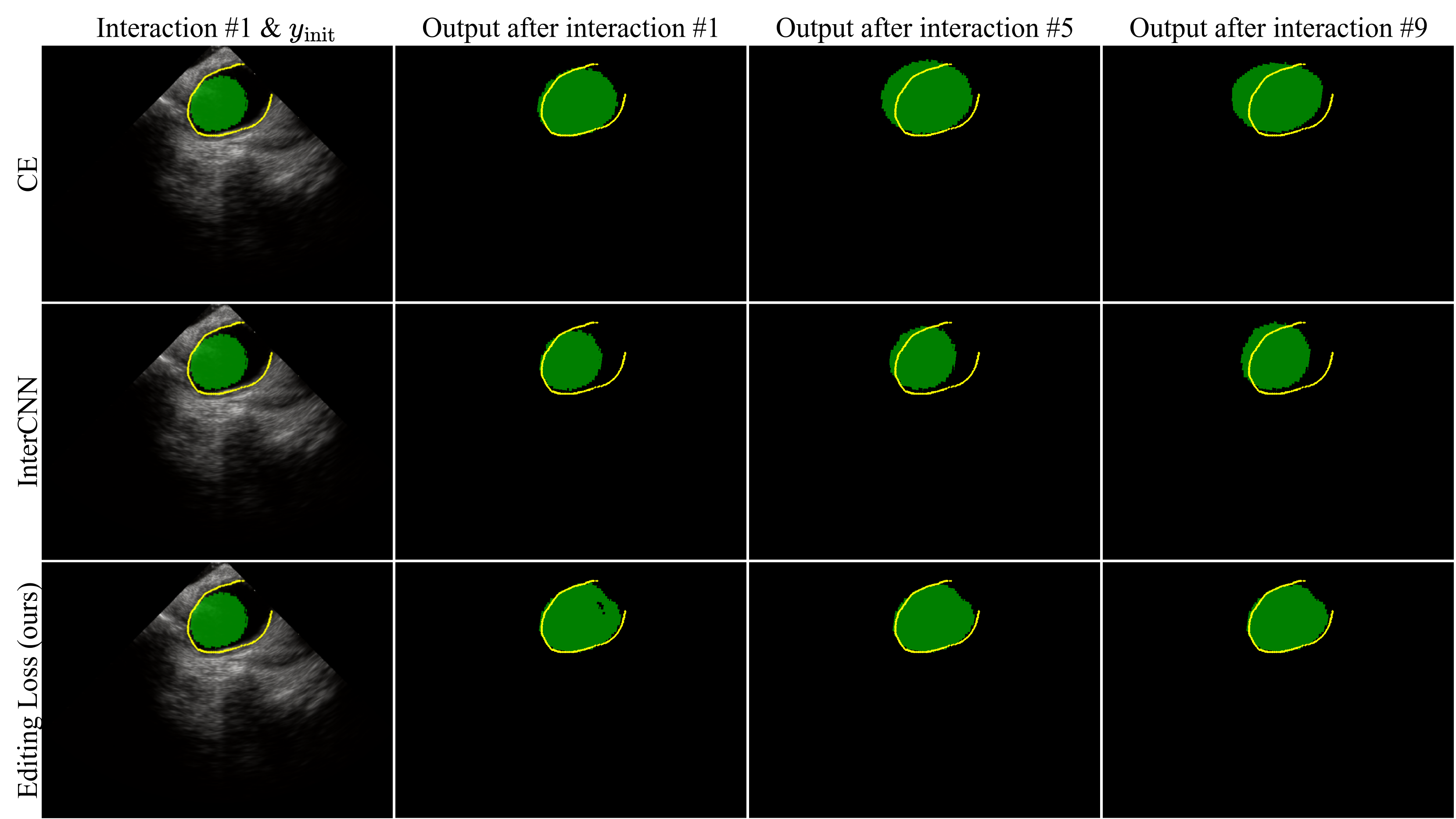}
    \caption{Impact of follow-up edits on the first correction. Yellow: first user edit. Green: output after each edit. Our editing loss maintains the integrity of the previous edits, while in the other methods the previous edits are compromised.}
    \label{fig:seq_editing_2}
\end{figure}

Furthermore, in Fig. \ref{fig:seq_editing_2}, we present a qualitative example to understand the effect of follow-up edits on the first correction. Edits after the first one are on other frames and not shown in the figure. We observe that the CE and InterCNN methods did not preserve the first correction, while the editing loss model maintained it. This is a crucial practical advantage of our loss, which allows the user to make corrections without compromising the previous edits.


\section{Conclusion}
We presented an interactive editing framework for challenging clinical applications. We devised an editing-specific loss function that penalizes the deviation from the ground truth near user interaction and penalizes deviation from the initial segmentation away from user interaction. Our novel editing algorithm is more robust as it does not compromise previously corrected regions. We demonstrate the performance of our method on the challenging task of volumetric segmentation of sparse \ice data. However, our formulation can be applied to other editing tasks and different imaging modalities.

\bibliographystyle{splncs04}
\bibliography{refs}

\clearpage
\title{Supplementary Material: From Sparse to Precise: A Practical Editing Approach for Intracardiac Echocardiography Segmentation}
\titlerunning{A Practical Editing Approach for ICE Segmentation}
\author{}
\institute{}
\maketitle              %
\begin{table}[!h]
\centering
\caption{Hyperparameters used for the reported experiments. The hyperparameters were tuned based on cross-validation performance. The value used for the number of steps in the InterCNN method was chosen based on the authors' recommendation, higher values increase training time significantly. All experiments performed on NVIDIA V100 GPU with 16 GB memory.}
\begin{tabular}{@{}lcc@{}}
\toprule
Hyperparameter & Values Explored & Optimal Value \\ \midrule
Learning Rate & 0.01, 0.001, 0.005, 0.0001 & 0.005 \\
Batch Size & 4 & 4 \\
Number of Filters per UNet Level & \{8, 16, 32, 64\}, \{16, 32, 64, 128\} & \{8, 16, 32, 64\} \\
$\sigma_{edit}$ & 10, 20, 30, 40, 50 & 20 \\
$\sigma_{enc}$ & 20 & 20 \\
InterCNN Steps & 10 & 10 \\ \bottomrule
\end{tabular}
\end{table}

\begin{figure}
    \centering
    \includegraphics[width=\textwidth]{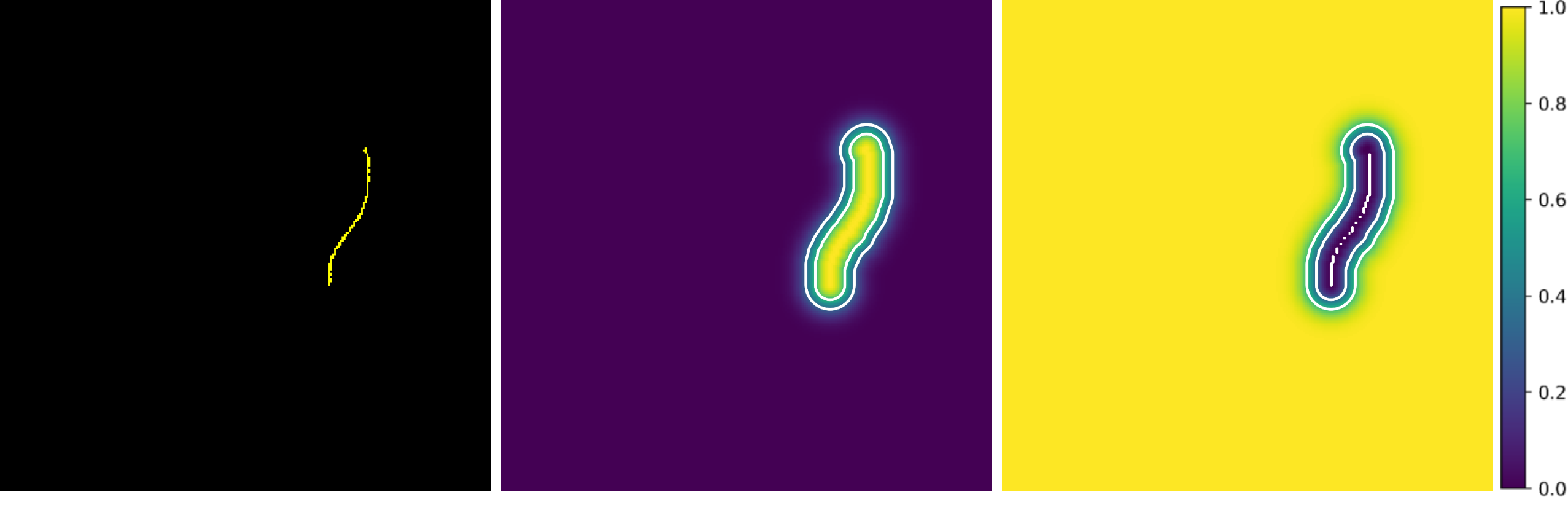}
    \caption{Visualization of an example user edit (left) and the corresponding $A$ and $\bar{A}$ matrices (middle and right, respectively). $A$ defines the vicinity of the user interaction, while $\bar{A}$ defines the regions far from it. Note that the presented visualization is in 2D for clarity, but in our experiments, we used 3D Gaussians to represent $A$ and $\bar{A}$ in all three dimensions of the ICE volume.}
\end{figure}

\begin{table}[!ht]
\centering
\caption{Proposed Future Work to Improve Segmentation Accuracy: Investigating Adaptive $\sigma_{edit}$ Values Based on Error Region Area}
\begin{tabular}{@{}p{0.9\linewidth}@{}}
\toprule
\textbf{Limitations and Future Work} \\
\midrule
\textbf{Limitation:} If the error to be corrected is larger than the interaction domain of influence defined by $A$, the updated segmentation may contain disconnected components.\\
\textbf{Future Work:} Investigate the use of adaptive $\sigma_{edit}$ values based on the area of the error region. This approach has the potential to enhance the overall accuracy of the segmentation process and reduce the occurrence of disconnected components. \\
\midrule
\textbf{Limitation:} The existing framework presumes user interaction occurs precisely at the boundary of the region as indicated by ground truth data, without considering potential inaccuracies arising from human error or noise. \\
\textbf{Future Work:} Rigorous experimental validation is recommended to examine the resilience of the method in response to perturbations in user input. This evaluation will allow us to ascertain the robustness of the methodology in real-world settings, where input may be subject to variability. \\
\bottomrule
\end{tabular}
\end{table}

\begin{algorithm}
\caption{Computation of the Editing Loss Function}
\begin{algorithmic}[1]
\State \textbf{Input:} $y$, $\hat{y}$, $y_{init}$, $u$, $\sigma_{edit}$
\State $A \gets \text{make\_3d\_gaussian\_map}(u, \sigma_{edit})$ \Comment{Create 3D Gaussian map}
\State $\mathcal{L}_{\text{edit}} \gets  - A [y\log(\hat{y}) + (1 - y)\log(1 - \hat{y})]$
\State $\mathcal{L}_{\text{preserve}} \gets - (1 - A)[y_{init}\log(\hat{y}) + (1 - y_{init}) \log(1 - \hat{y})]$
\State $\mathcal{L} \gets \mathcal{L}_{edit} + \mathcal{L}_{preserve}$
\State \textbf{Output:} $\frac{1}{|\mathcal{L}|} \sum \mathcal{L}$ \Comment{Return mean of the loss}
\end{algorithmic}
\end{algorithm}

\end{document}